\documentclass[11pt]{article}

\usepackage{acl}

\usepackage{times}
\usepackage{latexsym}

\usepackage[T1]{fontenc}

\usepackage[utf8]{inputenc}

\usepackage{microtype}

\usepackage{inconsolata}

\usepackage[most]{tcolorbox}
\usepackage{enumitem}
\usepackage{amssymb}
\usepackage{booktabs}
\usepackage{multirow}
\usepackage{makecell}
\usepackage{graphicx}
\usepackage{siunitx}
\usepackage{float}
\usepackage{xcolor}
\usepackage{hyperref}

\title{INS-ActBench: A Comprehensive Benchmark for Assessing Professional Actuarial Capability of Large Language Models}

\author{
  \textbf{Changyu Chen},
  \textbf{Chenwei Lin}\thanks{Project lead.},
  \textbf{Xian Xu}\thanks{Corresponding author.}
\\
  Fudan University
\\
  \small{
    \textbf{Correspondence:} \href{mailto:xianxu@fudan.edu.cn}{xianxu@fudan.edu.cn}
  }
}

\begin{document}
\maketitle
\begin{abstract}

Large Language Models (LLMs) have shown strong potential in financial reasoning, but existing benchmarks often evaluate domain knowledge, numerical reasoning, long-context understanding, and tool use in separate settings. This limits their ability to assess realistic professional workflows that require auditable, context-grounded, and tool-executable decisions. We introduce \textbf{INS-ActBench}, a comprehensive benchmark for evaluating professional actuarial capability in LLMs. INS-ActBench contains 12,050 Q\&A pairs from public exams and sample questions released by 16 actuarial associations. It covers three subsets: \textbf{INS-Act-Know} for standardized actuarial knowledge, \textbf{INS-Act-Case} for long-context insurance case reasoning, and \textbf{INS-Act-Practice} for spreadsheet and R-code tasks with verifiable numerical outputs. Experiments on nine representative LLMs and human actuarial experts reveal a clear capability boundary: frontier LLMs perform strongly on standardized knowledge, but remain much weaker in case reasoning, tool-based workflows, and jurisdiction-sensitive practice. INS-ActBench provides a reproducible foundation for developing actuarial LLMs toward reliable professional assistance. The code is available at \url{https://github.com/FDU-INS/INS-ActBench}.

\end{abstract}








\section{Introduction}

\begin{figure*}[t]
  \centering
  \includegraphics[width=0.95\textwidth]{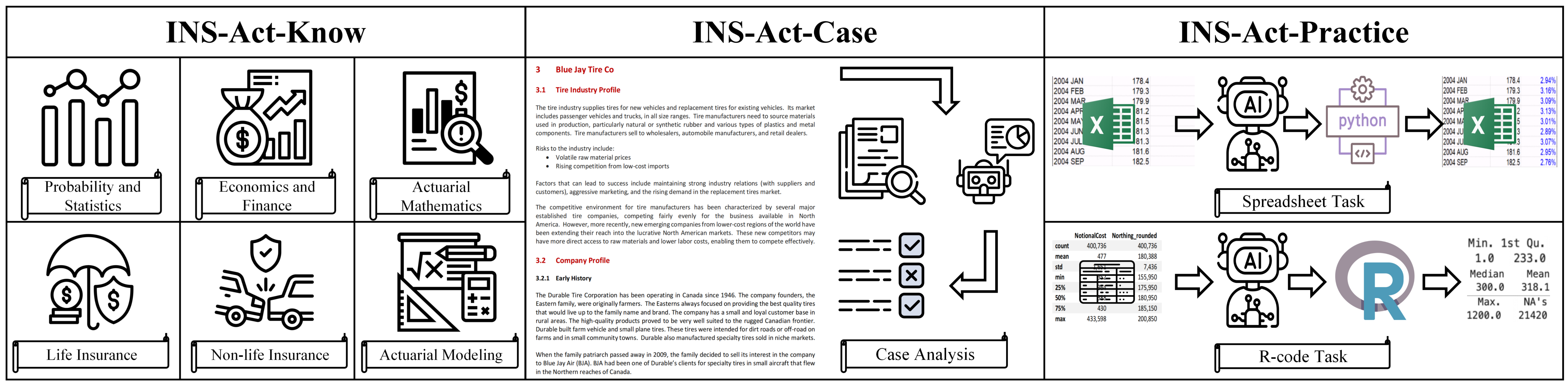}
  \vspace{-1mm}
  \caption{Evaluation dataset for INS-ActBench.}
  \label{fig:INS_ActBench}
  \vspace{-2mm}
\end{figure*}

Large Language Models (LLMs) have made rapid progress in natural language processing and are increasingly being applied to specialized professional domains that require expert knowledge, complex reasoning, and reliable decision support \citep{xie2024finben}. Finance is one such high-stakes domain. Financial professionals must interpret domain-specific documents, perform numerical analysis, assess risk, and make decisions under uncertainty \citep{xie2023pixiu,zhao2022multihiertt,li2025investorbench,gan2024mmefinance,tang2025financereasoning}. Consequently, financial benchmarks have become an important testbed for examining whether LLMs can move beyond general language understanding toward professional analytical capability.

However, existing benchmarks largely evaluate the components of professional capability in isolation. Prior work has examined financial text understanding \citep{xie2023pixiu}, question answering \citep{xie2024finben}, investment decision-making \citep{li2025investorbench}, table-text reasoning \citep{zhu2021tat}, long-context document understanding \citep{reddy2024docfinqa}, spreadsheet reasoning \citep{ravnik2026finsheet}, and code generation \citep{song2025statllm}. Although these benchmarks provide valuable evidence about individual model capabilities, real professional work rarely consists of a single isolated task. Instead, it requires models to connect domain knowledge, contextual evidence, regulatory assumptions, numerical reasoning, and tool-based implementation within a coherent and auditable workflow. In high-stakes settings, producing a plausible answer is therefore insufficient; professional assistance must also be context-grounded, computationally reproducible, and numerically verifiable.

Actuarial science provides a particularly rigorous setting for evaluating this integrated professional capability. As an interdisciplinary professional field, actuarial science combines mathematics, statistics, economics, finance, and insurance to quantify and manage the financial consequences of risk and uncertainty \citep{miljkovic2017computational}. Actuarial work supports pricing, reserving, capital management, solvency assessment, and long-term financial forecasting, and is therefore both calculation-intensive and highly dependent on business and regulatory context \citep{acharya2009financial,espinosa2021importance,owadally2018insurance}. Moreover, actuarial analyses are commonly implemented through spreadsheets and statistical programming environments such as R \citep{campbell2010spreadsheet,dutang2008actuar}. These characteristics make actuarial work especially suitable for testing whether LLMs can transform domain knowledge into context-grounded reasoning and executable, verifiable professional workflows. However, existing actuarial LLM studies remain limited to individual applications \citep{balona2024actuarygpt}, while insurance and financial benchmarks typically evaluate knowledge, retrieval, numerical reasoning, long-context understanding, or tool use separately \citep{zhang2025insqabench,chen2025inseva,lin2025insmmbench,zhou2025design}. It remains unclear whether current LLMs can progress from actuarial knowledge to case-based reasoning and ultimately to verifiable professional execution.

To address this gap, we introduce \textbf{INS-ActBench}, a comprehensive benchmark for evaluating the professional actuarial capabilities of LLMs. INS-ActBench comprises 12,050 question--answer pairs constructed from actuarial examination materials issued by 16 actuarial associations through a systematic pipeline of question selection, task restructuring, format standardization, annotation, and expert verification. Rather than treating actuarial evaluation as a collection of unrelated tasks, INS-ActBench follows a progressive capability structure. \textbf{INS-Act-Know} evaluates standardized actuarial knowledge across six subcategories. \textbf{INS-Act-Case} evaluates long-context reasoning over realistic insurance business cases. \textbf{INS-Act-Practice} evaluates whether models can translate actuarial requirements into executable spreadsheet and R-code workflows with numerically verifiable outputs. We further annotate source associations, jurisdiction-related contexts, and numerical question types to support fine-grained analysis of model performance.

Experiments on nine representative LLMs and an actuarially trained human baseline reveal a clear gap between actuarial knowledge and professional work competence. Frontier LLMs perform strongly on standardized actuarial knowledge and can exceed the sampled human baseline on this subset. Their performance, however, declines substantially on long-context case reasoning and practice-oriented tool use, where human participants remain more consistent. Additional analyses reveal substantial variation across numerical task types, workflow stages, and source associations. These findings suggest that current LLMs are increasingly capable of standardized actuarial calculation and knowledge-based reasoning, but reliable actuarial assistance still requires stronger contextual integration, tool-based execution, and consistency across diverse professional settings.

Our contributions are summarized below:
\begin{itemize}[leftmargin=1.2em,itemsep=0pt,topsep=2pt,parsep=0pt,partopsep=0pt]
    \item We introduce \textbf{INS-ActBench}, the first large-scale actuarial benchmark spanning knowledge, case reasoning, and tool-based practice.
    \item We develop a systematic pipeline that transforms actuarial examination materials from 16 associations into 12,050 standardized benchmark instances spanning knowledge, case reasoning, and professional execution.
    \item We compare representative LLMs with an actuarially trained human baseline, revealing a clear gap between standardized actuarial performance and reliable professional workflow execution.
\end{itemize}

\section{Related Work}

\begin{table*}[t]
\centering
\small
\setlength{\tabcolsep}{5pt}
\renewcommand{\arraystretch}{1.12}
\newcommand{\cmark}{\checkmark}
\newcommand{\xmark}{$\times$}
\resizebox{\textwidth}{!}{%
\begin{tabular}{@{}llccccc@{}}
\toprule
\textbf{Benchmark} & \textbf{Task Type}
& \textbf{Insurance}
& \textbf{Numerical}
& \textbf{Long Ctx.}
& \textbf{Spreadsheet}
& \textbf{Code} \\
\midrule

FinBen~\citep{xie2024finben}
& QA and reasoning
& \xmark & \cmark & \xmark & \xmark & \xmark \\

CFinBench~\citep{nie2024cfinbench}
& Chinese evaluation
& \xmark & \cmark & \xmark & \xmark & \xmark \\

InvestorBench~\citep{li2025investorbench}
& Investment
& \xmark & \cmark & \xmark & \xmark & \xmark \\

MME-Finance~\citep{gan2024mmefinance}
& Multimodal
& \xmark & \cmark & \xmark & \xmark & \xmark \\

\midrule
InsQABench~\citep{zhang2025insqabench}
& Insurance Q\&A
& \cmark & \xmark & \xmark & \xmark & \xmark \\

INSEva~\citep{chen2025inseva}
& Insurance evaluation
& \cmark & \xmark & \xmark & \xmark & \xmark \\

INS-MMBench~\citep{lin2025insmmbench}
& Insurance multimodal
& \cmark & \xmark & \xmark & \xmark & \xmark \\

\midrule
TAT-QA~\citep{zhu2021tat}
& Table-text reasoning
& \xmark & \cmark & \xmark & \xmark & \xmark \\

FinMathBench~\citep{he2026finmathbench}
& Math reasoning
& \xmark & \cmark & \xmark & \xmark & \xmark \\

\midrule
FinanceBench~\citep{islam2023financebench}
& Document Q\&A
& \xmark & \xmark & \cmark & \xmark & \xmark \\

DocFinQA~\citep{reddy2024docfinqa}
& Long-ctx. reasoning
& \xmark & \cmark & \cmark & \xmark & \xmark \\

\midrule
FinSheet-Bench~\citep{ravnik2026finsheet}
& Spreadsheets
& \xmark & \cmark & \xmark & \cmark & \xmark \\

Finch~\citep{dong2025finch}
& Workflows
& \xmark & \cmark & \xmark & \cmark & \cmark \\

\midrule
StatLLM~\citep{song2025statllm}
& Statistical coding
& \xmark & \cmark & \xmark & \xmark & \cmark \\

\midrule
\textbf{INS-ActBench (ours)}
& \textbf{Actuarial tasks}
& \cmark & \cmark & \cmark & \cmark & \cmark \\
\bottomrule
\end{tabular}%
}
\vspace{-1mm}
\caption{Comparison with existing financial benchmarks based on insurance area, numerical calculation, long-context, spreadsheet, coding. INS-ActBench includes all of them.}
\label{tab:related_benchmark_comparison}
\vspace{-2mm}  
\end{table*}

Existing financial benchmarks can be broadly grouped into three categories. First, general-purpose and domain-oriented benchmarks evaluate financial knowledge, question answering, decision-making, and multimodal understanding. Representative examples include PIXIU, FinBen, CFinBench, InvestorBench, and MME-Finance~\citep{xie2023pixiu,xie2024finben,nie2024cfinbench,li2025investorbench,gan2024mmefinance}. Insurance-oriented benchmarks further extend evaluation to insurance knowledge, document-based question answering, and multimodal understanding~\citep{zhang2025insqabench,chen2025inseva}. Second, numerical and document-reasoning benchmarks evaluate table-text reasoning, financial mathematics, and long-context question answering, including TAT-QA, FinMathBench, FinanceBench, and DocFinQA~\citep{zhu2021tat,he2026finmathbench,islam2023financebench,reddy2024docfinqa}. Third, tool-oriented benchmarks assess spreadsheet-centered workflows or statistical code generation and execution, such as FinSheet-Bench, Finch, and StatLLM~\citep{ravnik2026finsheet,dong2025finch,song2025statllm}. Table~\ref{tab:related_benchmark_comparison} summarizes the capability coverage of these benchmarks.

\begin{figure*}[t]
  \centering
  \includegraphics[width=0.9\textwidth]{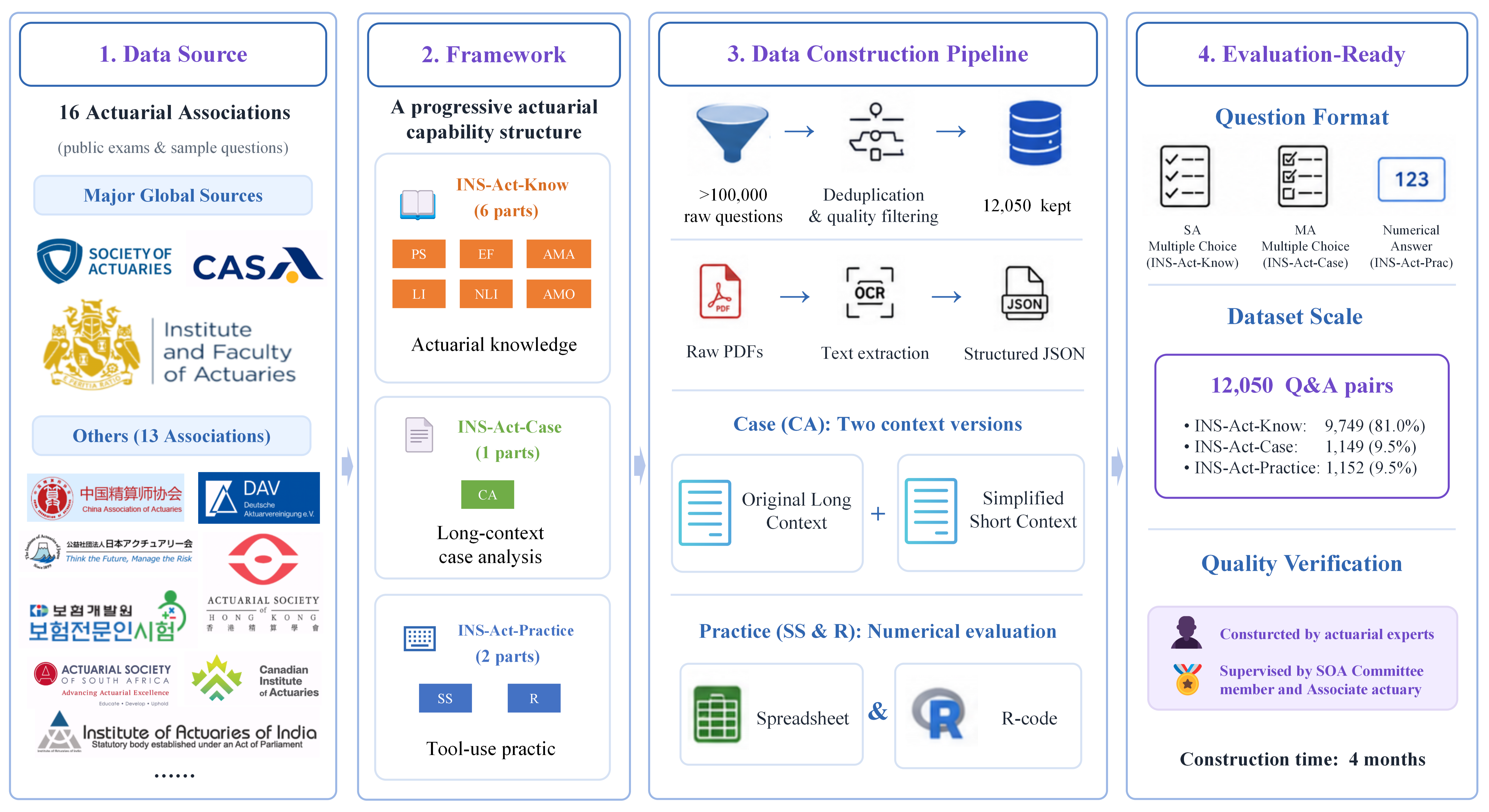}
  \vspace{-1mm}
  \caption{Overview of INS-ActBench pipeline.}
  \label{fig:ins_actbench_pipeline}
  \vspace{-2mm}
\end{figure*}

Although these benchmarks cover important components of financial reasoning, they generally evaluate them as separate capabilities. Actuarial work, by contrast, requires these capabilities to be integrated within a single professional workflow: models must interpret long business contexts, apply actuarial and regulatory assumptions, perform numerical calculations, and translate their reasoning into executable spreadsheet or code-based procedures~\citep{dutang2008actuar,acharya2009financial}. Existing financial and insurance benchmarks do not systematically evaluate this progression from domain knowledge to case-based reasoning and verifiable professional execution. INS-ActBench addresses this gap by jointly evaluating actuarial knowledge, long-context case analysis, and tool-oriented practice in a unified framework.


\section{INS-ActBench}

Figure~\ref{fig:ins_actbench_pipeline} presents the overall construction pipeline of INS-ActBench. We begin with professionally grounded actuarial examination materials from multiple associations and use them to define the scope of the benchmark. We then organize actuarial capability into a progressive framework spanning standardized knowledge, case-based reasoning, and tool-oriented practice. Based on this framework, we construct benchmark instances through systematic selection, task transformation, format standardization, and annotation, followed by multi-stage quality verification. The following subsections describe the source coverage, capability framework, benchmark construction process, and quality assurance procedure in sequence.
\subsection{Source Materials and Coverage}

INS-ActBench is systematically constructed on the basis of actuarial examination materials from 16 professional associations. We use the Society of Actuaries (SOA) and the Institute and Faculty of Actuaries (IFoA) as the two major global source categories, as they are among the largest actuarial credentialing systems worldwide; details are provided in Appendix~\ref{sec:appendixA}. The Casualty Actuarial Society (CAS) is grouped with SOA because both are primarily associated with the North American actuarial system, and we refer to this combined category as `SOA' throughout the paper. Materials from the remaining 13 associations are grouped as `Others', with their full names listed in Appendix~\ref{sec:appendixA}. These source categories capture differences in actuarial education, professional standards, and jurisdiction-related contexts, enabling INS-ActBench to evaluate model performance across major global actuarial systems and broader local settings.

\begin{table*}[t]
\centering
\small
\setlength{\tabcolsep}{4pt}
\renewcommand{\arraystretch}{1.12}
\resizebox{\textwidth}{!}{%
\begin{tabular}{l*{12}{c}}
\toprule
\multirow{2}{*}{\textbf{Annotation}} 
& \multicolumn{7}{c}{\textbf{INS-Act-Know}} 
& \multicolumn{1}{c}{\textbf{INS-Act-Case}} 
& \multicolumn{3}{c}{\textbf{INS-Act-Practice}} 
& \multirow{2}{*}{\textbf{Total}} \\
\cmidrule(lr){2-8} \cmidrule(lr){9-9} \cmidrule(lr){10-12}
& \textbf{PS} 
& \textbf{EF} 
& \textbf{AMA} 
& \textbf{LI} 
& \textbf{NLI} 
& \textbf{AMO} 
& \textbf{Total} 
& \textbf{CA} 
& \textbf{SS} 
& \textbf{R} 
& \textbf{Total} 
& \\
\midrule
SOA    & 658  & 493  & 61   & 163  & 101 & 510 & 1986 & 560 & 146 & 36  & 182 & 2728 \\
IFoA   & 214  & 438  & 146  & 205  & 265 & 246 & 1514 & 316 & 83  & 216 & 299 & 2129 \\
Others & 1292 & 1094 & 927  & 1481 & 802 & 653 & 6249 & 273 & 407 & 264 & 671 & 7193 \\
\midrule
Num 
       & 2108 & 1309 & 1058 & 1735 & 1018 & 1317 & 8545 & -- & -- & -- & -- & -- \\
Non-num 
       & 56   & 716  & 76   & 114  & 150  & 92   & 1204 & -- & -- & -- & -- & -- \\
\midrule
\textbf{Total} 
       & \textbf{2164} 
       & \textbf{2025} 
       & \textbf{1134} 
       & \textbf{1849} 
       & \textbf{1168} 
       & \textbf{1409} 
       & \textbf{9749} 
       & \textbf{1149} 
       & \textbf{636} 
       & \textbf{516} 
       & \textbf{1152} 
       & \textbf{12050} \\
\bottomrule
\end{tabular}%
}
\vspace{-1mm}
\caption{Basic statistics of INS-ActBench.}
\label{tab:data_source_distribution}
\vspace{-2mm}
\end{table*}

\subsection{Task Framework}
\label{Section 3.1}

Based on an analysis of actuarial examination materials from 16 professional associations, we formalize actuarial capability as a three-level progression. The first level concerns standardized knowledge, including actuarial concepts, principles, formulas, and routine calculations. The second concerns case-based reasoning, requiring the integration of actuarial knowledge with business background, financial information, and contextual evidence. The third concerns professional execution, where actuarial assumptions and task requirements must be translated into verifiable computational workflows. Accordingly, INS-ActBench consists of three progressively structured subsets: INS-Act-Know, INS-Act-Case, and INS-Act-Practice.

\textbf{INS-Act-Know} evaluates standardized actuarial knowledge, including actuarial concepts, principles, formulas, computation, and modeling. It contains six subcategories: Probability and Statistics (PS), Economics and Finance (EF), Actuarial Mathematics (AMA), Life Insurance (LI), Non-life Insurance (NLI), and Actuarial Modeling (AMO).

\textbf{INS-Act-Case} evaluates long-context actuarial reasoning in realistic insurance business scenarios. It uses company background, financial statements, operating information, and related case materials to test whether models can locate relevant evidence, integrate scattered information, and make actuarial judgments under complex business conditions. It contains one subcategory: Case Analysis (CA).

\textbf{INS-Act-Practice} evaluates practice-oriented actuarial tool use. It requires LLMs to translate actuarial task requirements into executable spreadsheet formulas or R code, and to produce verifiable numerical results. It contains two subcategories: Spreadsheet (SS) and R-code (R).

\subsection{Data Construction}

\subsubsection{Question Selection}

We initially assembled more than 100,000 raw items from actuarial examination materials issued by 16 professional associations. Given the high similarity among some historical exam items and balancing the scale of the benchmark, we applied a selective filtering process. First, we prioritized questions from the most recent five years for each association to improve freshness. Second, we focused on numerical and calculation-oriented actuarial questions, while retaining only a small number of representative non-numerical questions that test core actuarial concepts. In addition, we observed that actuarial knowledge, case analysis, and tool-use tasks appear in credentialing exams at an approximate ratio of 8:1:1, so we followed this structure when determining the relative sizes of the subsets. This process resulted in 12,050 selected questions. The original materials were primarily in PDF format, which we processed through text-layer extraction or OCR and organized into a standardized JSON format.

\subsubsection{Task Standardization}
\label{sec: Task Standardization}

To support standardized and reproducible evaluation, we further adapted the selected questions into task formats aligned with the three subsets.

For INS-Act-Know, we standardize questions into single-answer multiple-choice format. Subjective questions are adapted into independent objective items while preserving the information required for deriving the answer. For questions that already meet our format requirements, we still revise the wording and shuffle the options while preserving the original meaning. We further annotate each question as numerical(Num) or non-numerical(Non-num) to distinguish calculation actuarial knowledge from concept knowledge. 

For INS-Act-Case, we convert subjective case questions into multiple-answer multiple-choice questions. This partially reduces the openness of the original answers, but makes the tasks more suitable for benchmark construction and automated evaluation. We preserve the original case materials as long-context inputs with a maximum length of 175K tokens and an average length of 58K tokens. Each case question is paired with an original long context and a simplified short context (reducing the context length to within 32K tokens) to support LLMs with different context lengths. The token distribution is shown in Figure~\ref{fig:context-token-distribution} in Appendix~\ref{sec:appendixB}. 

For INS-Act-Practice, we construct spreadsheet and R-code tasks. To ensure evaluation accuracy, we adapt all questions to require numerical answers. Each spreadsheet task contains an input spreadsheet for the LLM and a ground-truth spreadsheet as reference answer, which differ only in the answer cells for evaluation. We convert every cell in the input spreadsheet into textual input to avoid losing actuarial information. We require the LLM to write Python code that fills the predicted answers into predefined answer cells and regenerates the spreadsheet. Each R-code task require numerical outputs such as statistics, correlation coefficients, or model-selection criteria, with the original data provided in textual form. We require LLMs to generate R code that computes the corresponding numerical results.

This task standardization process adapts all selected questions into benchmark formats and reduces the risk of data contamination. We further conduct dedicated data contamination diagnostics in Appendix~\ref{app:contamination}, showing that potential contamination has limited impact on our results. Representative examples for each subset are provided in Appendix~\ref{sec:appendixB}. And Table~\ref{tab:data_source_distribution} provides a basic statistics overview of INS-ActBench.

\subsection{Quality Verification}

INS-ActBench was constructed and independently verified by experts with backgrounds in actuarial science and computer science. Ambiguous or disputed instances were further reviewed and calibrated by a senior actuarial expert who is a member of the SOA China Committee and an Associate of the SOA. In total, 9.65\% of the benchmark instances were revised during quality control, and the complete construction and verification process took four months. Further details are provided in Appendix~\ref{sec:appendixB}.

We also conducted data-exposure diagnostics to examine whether the results could be substantially driven by direct memorization of the source materials. Specifically, we compared model performance across items with different levels of online exposure, tested whether models could reconstruct the original answer options, and evaluated performance on recently released examination items. The results provide no clear evidence that direct exposure or option memorization explains the strong performance on INS-Act-Know. Detailed diagnostic results are reported in Appendices~\ref{app:contamination}.

\section{Experiments}

\subsection{Experimental Setup}
\label{sec: Experimental Setup}

\paragraph{Selected LLMs.}
We evaluate nine representative LLMs, including six proprietary models: GPT-5.5, Claude-Opus-4.7, Gemini-3.1-Pro, DeepSeek-V4-Pro, Kimi-K2.6, and Qwen3.6-Plus; and three open-source models: Qwen3.5-35B-A3B, Qwen3-14B, and Gemma-3-12B-IT. The selected models cover both frontier proprietary systems and widely used open-source model families.

\paragraph{Inference Settings.}
INS-Act-Know and INS-Act-Case use a two-shot setting, while INS-Act-Practice use a zero-shot setting. Proprietary models are run with thinking mode enabled and set to high when supported. Open-source models are are evaluated using the vLLM framework \citep{kwon2023vllm}. Due to context-length limits, Qwen3-14B and Gemma-3-12B-IT use simplified case contexts, while the other models use original long-context cases in INS-Act-Case. The system prompts are provided in Figure~\ref{fig:system_prompts} in Appendix~\ref{sec:appendixB}.

\paragraph{Human Baseline.}
We conduct a human expert evaluation on a sampled subset of INS-ActBench. Five actuarial participants completed the test independently: one expert holding a PhD degree in actuarial science, three experts holding a master’s degree in actuarial science, and one outstanding undergraduate student majoring in actuarial science. None of them participated in dataset construction. We sampled 300 questions, including 100 questions from each subset. Participants used offline computers equipped with Excel, R, and comprehensive electronic actuarial textbooks. Internet access, web search, and LLM-based assistance were disabled during the evaluation. Each participant spent approximately 20 hours completing the human evaluation.

\paragraph{Evaluation.}
(1) For single-answer multiple-choice questions, we report accuracy. (2) For multiple-answer multiple-choice questions, an answer is counted as correct only when all selected options exactly match the reference answer. (3) For Spreadsheet tasks, we execute the generated Python code to produce an output file, and then compare the numerical values in the predefined answer cells with the reference answers. If the response is a formula in answer cell, we execute it to obtain the numerical result. If multiple answer cells are required, we take the average accuracy of each cell. A relative error tolerance of 0.2\% is allowed, which is acceptable in actuarial science. (4) For R-code tasks, the generated code is executed in a Docker environment, and the extracted numerical outputs are compared with the reference answers. A relative error tolerance of 0.2\% is applied to both spreadsheet and R-code tasks to accommodate minor numerical and implementation differences.

To further analyze the failure reasons in INS-Act-Practice, we conduct an error analysis for all incorrect Spreadsheet and R-code responses, and classify each error into three categories: (1) \textbf{Tool call failure} refers to cases where the model fails to transform the task into an executable tool operation, causing the workflow to remain unstarted, the output to be missing. This reflects a pure tool-use problem. (2) \textbf{Function usage error} refers to cases where the model has entered the tool operation process, but makes operational mistakes in data access, worksheet organization, reference construction, or function usage. This reflects a mixed problem involving both tool use and actuarial knowledge. (3) \textbf{Modeling and calculation error} refers to cases where the model successfully completes the tool execution and produces an evaluable result, but obtains an answer inconsistent with the reference due to errors in actuarial model selection, formula specification, or numerical derivation; this reflects a pure actuarial knowledge problem.

\subsection{Main Results}

\begin{table*}[t]
\centering
\scriptsize
\setlength{\tabcolsep}{4pt}
\renewcommand{\arraystretch}{1.15}
\newcommand{\best}[1]{\textbf{#1}}
\resizebox{\textwidth}{!}{%
\begin{tabular}{
@{}
c l
*{12}{c}
@{}
}
\toprule
\multicolumn{2}{c}{\multirow{2}{*}{\textbf{Model}}}
& \multicolumn{7}{c}{\textbf{INS-Act-Know}}
& \multicolumn{1}{c}{\textbf{INS-Act-Case}}
& \multicolumn{3}{c}{\textbf{INS-Act-Practice}}
& \multicolumn{1}{c}{\multirow{2}{*}{\textbf{Total}}} \\
\cmidrule(lr){3-9}
\cmidrule(lr){10-10}
\cmidrule(lr){11-13}
\multicolumn{2}{c}{}
& \textbf{PS}
& \textbf{EF}
& \textbf{AMA}
& \textbf{LI}
& \textbf{NLI}
& \textbf{AMO}
& \textbf{Total}
& \textbf{CA}
& \textbf{SS}
& \textbf{R}
& \textbf{Total}
& {} \\
\midrule

\multirow{6}{*}{\makecell[c]{\textbf{Proprietary}\\\textbf{LLMs}}}
& \textbf{GPT-5.5}
& 97.0 & \best{95.4} & 93.1 & 87.2 & 93.2 & 92.5 & 93.3
& 56.3
& 67.7 & 71.1 & 69.2
& 72.9 \\

& \textbf{Gemini-3.1-Pro}
& \best{97.5} & \best{95.4} & 94.8 & \best{90.6} & \best{94.0} & \best{94.0} & \best{94.5}
& 55.5
& 67.0 & 69.0 & 67.9
& 72.7 \\

& \textbf{DeepSeek-V4-Pro}
& 96.5 & 93.1 & 92.2 & 83.3 & 90.2 & 91.0 & 91.3
& 43.9
& 55.2 & 69.6 & 61.7
& 65.6 \\

& \textbf{Claude-Opus-4.7}
& 84.7 & 85.5 & 76.6 & 72.9 & 83.5 & 81.8 & 81.1
& 43.3
& 60.6 & 71.3 & 65.4
& 63.3 \\

& \textbf{Qwen3.6-Plus}
& 95.1 & 90.7 & 89.9 & 73.5 & 89.2 & 88.7 & 87.8
& 44.6
& 47.4 & 69.8 & 57.4
& 63.3 \\

& \textbf{Kimi-K2.6}
& 78.7 & 75.8 & 64.7 & 57.2 & 76.2 & 70.5 & 70.9
& 24.8
& 29.5 & 69.0 & 47.2
& 47.6 \\

\addlinespace[2pt]
\midrule
\addlinespace[2pt]

\multirow{3}{*}{\makecell[c]{\textbf{Open-source}\\\textbf{LLMs}}}
& \textbf{Qwen3.5-35B-A3B}
& 43.4 & 43.5 & 33.8 & 29.9 & 37.9 & 36.8 & 38.1
& 20.2
& 16.3 & 64.0 & 37.7
& 32.0 \\

& \textbf{Qwen3-14B}
& 42.2 & 44.3 & 38.2 & 34.6 & 40.9 & 41.0 & 40.4
& 21.0
& 2.7 & 54.8 & 26.0
& 29.1 \\

& \textbf{Gemma-3-12B-IT}
& 33.6 & 42.0 & 32.2 & 27.3 & 36.4 & 36.1 & 34.7
& 16.7
& 1.0 & 44.4 & 20.4
& 24.0 \\

\addlinespace[2pt]
\midrule
\addlinespace[2pt]

\multicolumn{2}{c}{\textbf{Human Experts Baseline}}
& 84.7 & 84.7 & \best{97.7} & 81.2 & 84.7 & 72.0 & 84.4
& \best{79.6}
& \best{79.1} & \best{90.7} & \best{84.9}
& \best{83.0} \\

\bottomrule
\end{tabular}%
}
\vspace{-1mm}
\caption{Evaluation results of different LLMs and human experts on each subset of INS-ActBench. The values represent average accuracy. For each subset, the Total score is computed over all questions within that subset. The final Total is the arithmetic mean of the three subset Total scores, preventing differences in subset size from disproportionately affecting the overall result. The best performance is in \textbf{bold}.}
\label{tab:llm_benchmark_scores}
\vspace{-2mm}
\end{table*}

\begin{table*}[t]
\centering
\scriptsize
\setlength{\tabcolsep}{3pt}
\renewcommand{\arraystretch}{1.12}

\resizebox{\textwidth}{!}{%
\begin{tabular}{llccccccccccccccc}
\toprule
\multicolumn{2}{c}{\multirow{2}{*}{\textbf{Model}}}
& \multicolumn{2}{c}{\textbf{PS}}
& \multicolumn{2}{c}{\textbf{EF}}
& \multicolumn{2}{c}{\textbf{AMA}}
& \multicolumn{2}{c}{\textbf{LI}}
& \multicolumn{2}{c}{\textbf{NLI}}
& \multicolumn{2}{c}{\textbf{AMO}}
& \multicolumn{3}{c}{\textbf{Total}} \\
\cmidrule(lr){3-4}
\cmidrule(lr){5-6}
\cmidrule(lr){7-8}
\cmidrule(lr){9-10}
\cmidrule(lr){11-12}
\cmidrule(lr){13-14}
\cmidrule(lr){15-17}
\multicolumn{2}{c}{}
& \textbf{Num} & \textbf{Non-num}
& \textbf{Num} & \textbf{Non-num}
& \textbf{Num} & \textbf{Non-num}
& \textbf{Num} & \textbf{Non-num}
& \textbf{Num} & \textbf{Non-num}
& \textbf{Num} & \textbf{Non-num}
& \textbf{Num} & \textbf{Non-num} & \textbf{Gap} \\
\midrule

\multirow{6}{*}{\makecell[c]{\textbf{Proprietary}\\\textbf{LLMs}}}
& \textbf{Gemini-3.1-Pro}
& \textbf{97.6} & \textbf{94.6}
& \textbf{96.0} & 94.4
& \textbf{95.1} & \textbf{90.8}
& \textbf{90.5} & \textbf{93.0}
& \textbf{94.0} & \textbf{94.0}
& \textbf{94.1} & \textbf{93.5}
& \textbf{94.6} & 93.9 & \textcolor{green!50!black}{+0.7} \\

& \textbf{GPT-5.5}
& 97.2 & 92.9
& 95.0 & \textbf{96.1}
& 93.3 & \textbf{90.8}
& 86.9 & 92.1
& 93.4 & 92.0
& 92.5 & 92.4
& 93.1 & \textbf{94.4} & \textcolor{red}{-1.3} \\

& \textbf{DeepSeek-V4-Pro}
& 96.6 & 92.9
& 93.6 & 92.2
& 92.5 & 88.2
& 83.1 & 87.7
& 90.7 & 87.3
& 91.1 & 89.1
& 91.3 & 90.7 & \textcolor{green!50!black}{+0.6} \\

& \textbf{Qwen3.6-Plus}
& 95.1 & 92.9
& 91.1 & 89.9
& 89.8 & \textbf{90.8}
& 73.3 & 77.2
& 88.8 & 92.0
& 88.9 & 85.9
& 87.7 & 88.9 & \textcolor{red}{-1.2} \\

& \textbf{Claude-Opus-4.7}
& 84.4 & 92.9
& 82.3 & 91.3
& 75.9 & 86.8
& 71.9 & 87.7
& 82.8 & 88.0
& 81.2 & 90.2
& 79.8 & 90.3 & \textcolor{red}{-10.5} \\

& \textbf{Kimi-K2.6}
& 78.7 & 82.1
& 72.4 & 82.0
& 63.6 & 80.3
& 56.9 & 60.5
& 75.3 & 82.0
& 69.2 & 90.2
& 69.6 & 80.5 & \textcolor{red}{-10.9} \\

\midrule

\multirow{3}{*}{\makecell[c]{\textbf{Open-source}\\\textbf{LLMs}}}
& \textbf{Qwen3-14B}
& 41.5 & 69.6
& 28.8 & 72.6
& 35.5 & 75.0
& 32.3 & 68.4
& 36.3 & 72.0
& 38.6 & 76.1
& 35.9 & 72.4 & \textcolor{red}{-36.5} \\

& \textbf{Qwen3.5-35B-A3B}
& 42.5 & 76.8
& 25.7 & 76.0
& 31.0 & 72.4
& 27.4 & 67.5
& 32.4 & 75.3
& 34.0 & 76.1
& 32.9 & 74.9 & \textcolor{red}{-42.0} \\

& \textbf{Gemma-3-12B-IT}
& 32.8 & 64.3
& 26.8 & 69.8
& 30.5 & 55.3
& 25.0 & 62.3
& 32.7 & 61.3
& 33.9 & 67.4
& 30.2 & 66.7 & \textcolor{red}{-36.5} \\

\bottomrule
\end{tabular}%
}
\vspace{-1mm}
\caption{Performance on numerical and non-numerical questions in INS-Act-Know.}
\label{tab:numerical_non_numerical}
\vspace{-2mm}
\end{table*}

\begin{table*}[t]
\centering
\scriptsize
\setlength{\tabcolsep}{4pt}
\renewcommand{\arraystretch}{1.15}
\resizebox{\textwidth}{!}{%
\begin{tabular}{
@{}
c l
*{12}{S[table-format=2.1]}
@{}
}
\toprule
\multicolumn{2}{c}{\multirow{2}{*}{\textbf{Model}}}
& \multicolumn{3}{c}{\textbf{INS-Act-Know}}
& \multicolumn{3}{c}{\textbf{INS-Act-Case}}
& \multicolumn{3}{c}{\textbf{INS-Act-Practice}}
& \multicolumn{3}{c}{\textbf{Total}} \\
\cmidrule(lr){3-5} \cmidrule(lr){6-8} \cmidrule(lr){9-11} \cmidrule(lr){12-14}
\multicolumn{2}{c}{}
& {\textbf{SOA}} & {\textbf{IFoA}} & {\textbf{Others}}
& {\textbf{SOA}} & {\textbf{IFoA}} & {\textbf{Others}}
& {\textbf{SOA}} & {\textbf{IFoA}} & {\textbf{Others}}
& {\textbf{SOA}} & {\textbf{IFoA}} & {\textbf{Others}} \\
\midrule

\multirow{6}{*}{\makecell[c]{\textbf{Proprietary}\\\textbf{LLMs}}}
& \textbf{GPT-5.5}
& 97.7 & 95.2 & 91.4
& {\bfseries 55.0} & 43.4 & 74.0
& 80.3 & {\bfseries 80.0} & {\bfseries 61.4}
& {\bfseries 87.8} & 85.4 & 87.9 \\

& \textbf{Claude-Opus-4.7}
& 85.1 & 91.7 & 77.2
& 36.4 & 32.3 & 70.3
& 79.0 & 77.1 & 56.5
& 74.7 & 80.9 & 75.0 \\

& \textbf{Gemini-3.1-Pro}
& {\bfseries 98.6} & {\bfseries 95.6} & {\bfseries 93.0}
& 42.3 & {\bfseries 57.6} & {\bfseries 80.2}
& {\bfseries 86.8} & 78.0 & 58.3
& 86.3 & {\bfseries 87.5} & {\bfseries 89.3} \\

& \textbf{DeepSeek-V4-Pro}
& 97.7 & 91.7 & 89.1
& 37.1 & 32.6 & 70.7
& 72.2 & 79.3 & 50.9
& 83.6 & 81.2 & 84.8 \\

& \textbf{Kimi-K2.6}
& 75.7 & 81.9 & 66.7
& 28.9 & 16.5 & 26.0
& 37.7 & 71.2 & 39.1
& 63.6 & 70.7 & 62.6 \\

& \textbf{Qwen3.6-Plus}
& 95.7 & 93.5 & 84.0
& 36.1 & 36.1 & 71.8
& 56.8 & 74.7 & 49.9
& 80.9 & 82.3 & 80.3 \\

\addlinespace[2pt]
\midrule
\addlinespace[2pt]

\multirow{3}{*}{\makecell[c]{\textbf{Open-source}\\\textbf{LLMs}}}
& \textbf{Qwen3.5-35B-A3B}
& 30.5 & 68.6 & 33.1
& 12.3 & 12.0 & 45.8
& 29.5 & 61.3 & 29.4
& 26.7 & 59.2 & 33.3 \\

& \textbf{Qwen3-14B}
& 32.4 & 72.0 & 35.3
& 18.9 & 11.4 & 36.3
& 20.9 & 50.0 & 16.7
& 28.9 & 59.9 & 33.6 \\

& \textbf{Gemma-3-12B-IT}
& 29.6 & 62.9 & 29.5
& 15.4 & 7.0 & 30.8
& 17.6 & 41.3 & 11.9
& 25.9 & 51.6 & 27.9 \\

\bottomrule
\end{tabular}%
}
\vspace{-1mm}
\caption{Cross-jurisdictional performance on INS-ActBench.}
\label{tab:llm_scores_by_Cross-Jurisdiction}
\vspace{-2mm}
\end{table*}

\textbf{Current LLMs show strong actuarial knowledge mastery but still lack reliable actuarial work execution.} Table~\ref{tab:llm_benchmark_scores} shows that proprietary LLMs perform strongly on standardized actuarial knowledge, but their scores decline sharply in case analysis and practice-oriented tool use. GPT-5.5 achieves the highest total score among the evaluated LLMs and Gemini-3.1-Pro achieves the highest INS-Act-Know score; Open-source models remain far behind, especially on CA and SS tasks. The results suggest that actuarial knowledge questions mainly test rule-based concepts and short-chain numerical reasoning, where frontier LLMs can rely on stable domain regularities and calculation schemas. Case and tool-use tasks require evidence localization, assumption tracking, cross-table computation, and executable workflow construction. INS-ActBench therefore identifies the boundary between actuarial knowledge mastery and actuarial work competence, providing a reproducible benchmark for developing LLMs that can support realistic actuarial workflows with verifiable outputs. This finding provides a practical roadmap for future actuarial LLM development: moving from strong exam-style knowledge performance toward trustworthy decision support across real insurance workflows.

\begin{figure}[t]
  \centering
  \includegraphics[width=0.95\columnwidth]{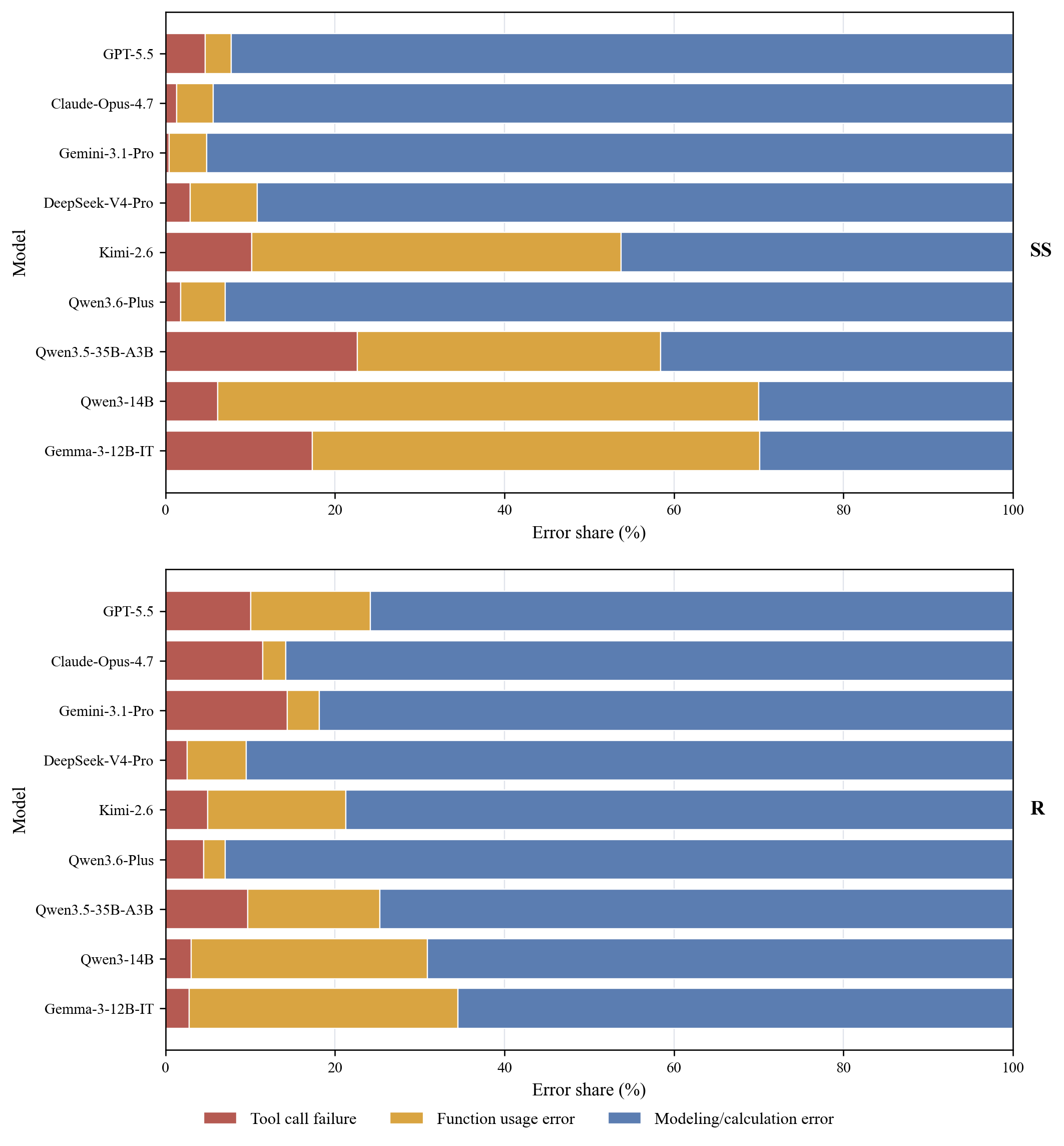}
  \caption{Error distributions for INS-Act-Practice. Top: spreadsheet tasks; bottom: R-code tasks.}
  \label{fig:error_analysis}
\end{figure}

\textbf{Compared with human experts, LLMs lead in standardized actuarial knowledge but lag behind in case reasoning and practical tool use.} Proprietary LLMs have exceeded actuarial experts on actuarial knowledge, but fall below on case and practice. Human experts are expected to remain stable across dimensions because actuarial training emphasizes applying principles under business context and practical calculation procedures. LLMs show more uneven results because their strengths are concentrated in standardized textual and numerical patterns, while long-context integration and executable tool-based workflows demand greater reliability. Recent studies show that although LLMs have achieved high accuracy on standard mathematical benchmarks, this capability does not directly translate into real-world applications\citep{cao2026contextmath}. Our finding indicates that current LLMs have already demonstrated strong capabilities in actuarial calculation and knowledge checking, and future actuarial LLM development should prioritize context-grounded reasoning, reliable tool execution, and expert-level consistency across the full actuarial workflow.  

To further interpret main results, we conduct three additional analyses in next three subsections. More experimental results are provided in Appendix~\ref{sec:appendixC}.

\subsection{Numerical versus Non-numerical Performance}

\textbf{Numerical actuarial reasoning distinguishes frontier models from weaker LLMs.} Table~\ref{tab:numerical_non_numerical} shows that frontier LLMs perform strongly on both numerical and non-numerical questions in INS-Act-Know based on Section~\ref{sec: Task Standardization}, with numerical questions even yielding higher scores in most categories. This suggests that leading LLMs can already handle formula-based actuarial computation when the problem provides standard models, and well-defined calculation steps. However, LLMs with weaker numerical reasoning ability show much lower accuracy on numerical questions, indicating that actuarial numerical reasoning remains a major capability barrier for many LLMs. This decomposition therefore helps INS-ActBench identify whether a model’s weakness comes from actuarial calculation, conceptual understanding, or both, and highlights actuarial numerical reasoning as an important direction for future LLM improvement.

\subsection{Practice Error Analysis}

\textbf{Failures in INS-Act-Practice are mainly concentrated in knowledge-related practice capabilities.} Figure~\ref{fig:error_analysis} shows that across both SS and R tasks, the tool call failures account for only a small share for LLMs. This pattern indicates that current LLMs can usually enter the tool-use workflow and produce executable outputs, but still struggle to transform spreadsheet/R operations into correct actuarial procedures. These results show that INS-Act-Practice evaluates a deeper form of actuarial competence: future actuarial LLMs need stronger integration between tool operation and actuarial reasoning, rather than only better tool invocation.

\subsection{Cross-Jurisdictional Performance}

\textbf{Actuarial reasoning by LLMs remains sensitive to regulatory context, especially when tasks move beyond standardized knowledge into case analysis and practice.} Table~\ref{tab:llm_scores_by_Cross-Jurisdiction} shows clear performance variation across SOA, IFoA, and Other sources, and this variation becomes more visible in case analysis and practice-oriented tasks than in standardized knowledge questions. This suggests that LLMs do not handle actuarial tasks uniformly across regulatory environments, even when the tasks belong to the same professional domain. The reason is that insurance is strongly shaped by jurisdiction-specific regulation, so actuarial reasoning often depends on local institutional context rather than general abstract technical knowledge alone. These results indicate that actuarial LLMs should be evaluated and improved under diverse global regulatory settings instead of being optimized for a single actuarial environment.

\section{Conclusion}

In this paper, we introduced INS-ActBench, a large-scale benchmark for evaluating LLMs in actuarial science. INS-ActBench covers actuarial knowledge, long-context case analysis, and practice-oriented tool use with spreadsheets and R code. Experiments on nine representative LLMs and human experts show that current models perform strongly on standardized actuarial calculations, yet still lag behind experts in long-context case reasoning and tool-based numerical workflows. These findings highlight actuarial science as a rigorous domain for testing financial reasoning and provide a benchmark for future actuarial LLM research.


\section*{Limitations}

INS-ActBench evaluates standardized core capabilities relevant to professional actuarial work, but it does not reproduce dynamic client interaction or iterative workplace updates. Some subjective actuarial questions are converted into objective formats for reproducible evaluation, which improves scoring consistency but may reduce the openness of real professional judgment. In addition, the current `Numerical’ annotation does not distinguish the type of actuarial modelling components, and future versions will extend the dataset with finer-grained difficulty annotations and broader interaction settings. Finally, although INS-ActBench covers major actuarial source categories and broader local associations, its results may still reflect the distribution of available public materials and may not represent every regulatory environment, language context, or firm-specific actuarial practice.

\section*{Ethical Considerations}

INS-ActBench is constructed from publicly available professional materials and is intended for research and educational evaluation. The benchmark does not contain private personal information, confidential policyholder records, or proprietary corporate data. Because actuarial work is closely related to insurance pricing, reserving, solvency assessment, and regulatory reporting, we emphasize that benchmark performance should not be interpreted as professional actuarial qualification or as a basis for real financial decisions. LLM outputs evaluated by INS-ActBench require review by certified actuarial professionals before any practical use. The dataset is designed to measure model capability and identify failure modes in actuarial reasoning, tool use, and jurisdiction-sensitive interpretation, and the authors do not assume responsibility for downstream decisions made using models evaluated on this benchmark.

\section*{Acknowledgements}

We used generative AI tools only for language polishing and short-form input assistance during the writing process. This assistance was limited to language-level refinement.


\bibliography{custom}

\clearpage

\appendix

\section{Global Actuarial Association}
\label{sec:appendixA}

According to the Members with Actuarial Credentials (MWACs) statistics released by the International Actuarial Association (IAA) in March 2026\footnote{\url{https://actuaries.org/committees/membership-assistance/membership-assistance-reference-materials/}}, SOA/IFoA/CAS are the three largest actuarial associations by reported credentialed members, together accounting for 57.84\% of global MWACs. This indicates that the North American actuarial system, represented by SOA and CAS, and the British actuarial system, represented by IFoA, are the two most influential global actuarial credentialing systems. The remaining 71 actuarial associations account for 42.16\% of MWACs and reflect diverse local regulatory, examination, and professional practice environments. Details are shown in Table~\ref{tab:iaa_mwac_2026_top3}. Therefore, INS-ActBench uses SOA/CAS and IFoA as the two major global sources, while further incorporating 13 other actuarial associations to capture broader jurisdiction-specific actuarial characteristics. They are:

China Association of Actuaries (CAA); Institute of Actuaries of Japan (IAJ); Institute of Actuaries of India (IAI); Persatuan Aktuaris Indonesia / Society of Actuaries of Indonesia (PAI); Institute of Actuaries of Korea (IAK); Deutsche Aktuarvereinigung / German Association of Actuaries (DAV); Instituto Brasileiro de Atuária / Brazilian Institute of Actuaries (IBA); Actuarial Society of Hong Kong (ASHK); Canadian Institute of Actuaries (CIA); Actuarial Society of South Africa (ASSA); Japanese Society of Certified Pension Actuaries (JSCPA); Actuaries Institute Australia (AIA); Israel Association of Actuaries (ISOA).

\begin{table}[H]
\centering
\small
\begin{tabular}{clrr}
\toprule
\textbf{No.} & \textbf{Association} & \textbf{Members} & \textbf{Share} \\
\midrule
1 & SOA    & 34,466  & 30.99\% \\
2 & IFoA   & 18,406  & 16.55\% \\
3 & CAS    & 11,451  & 10.30\% \\
4 & Others & 46,890  & 42.16\% \\
\midrule
-- & Total & 111,213 & 100.00\% \\
\bottomrule
\end{tabular}
\vspace{-1mm}
\caption{Top actuarial associations by reported MWACs.}
\label{tab:iaa_mwac_2026_top3}
\vspace{-2mm}
\end{table}

\section{Dataset Construction Details}
\label{sec:appendixB}

In filtering more than 100,000 raw questions down to the final 12,050 items, we applied several selection criteria, with the approximate proportions removed at each stage reported in Table~\ref{tab:selection_funnel}.

\begin{figure*}[t]
  \centering
  \includegraphics[width=0.95\textwidth]{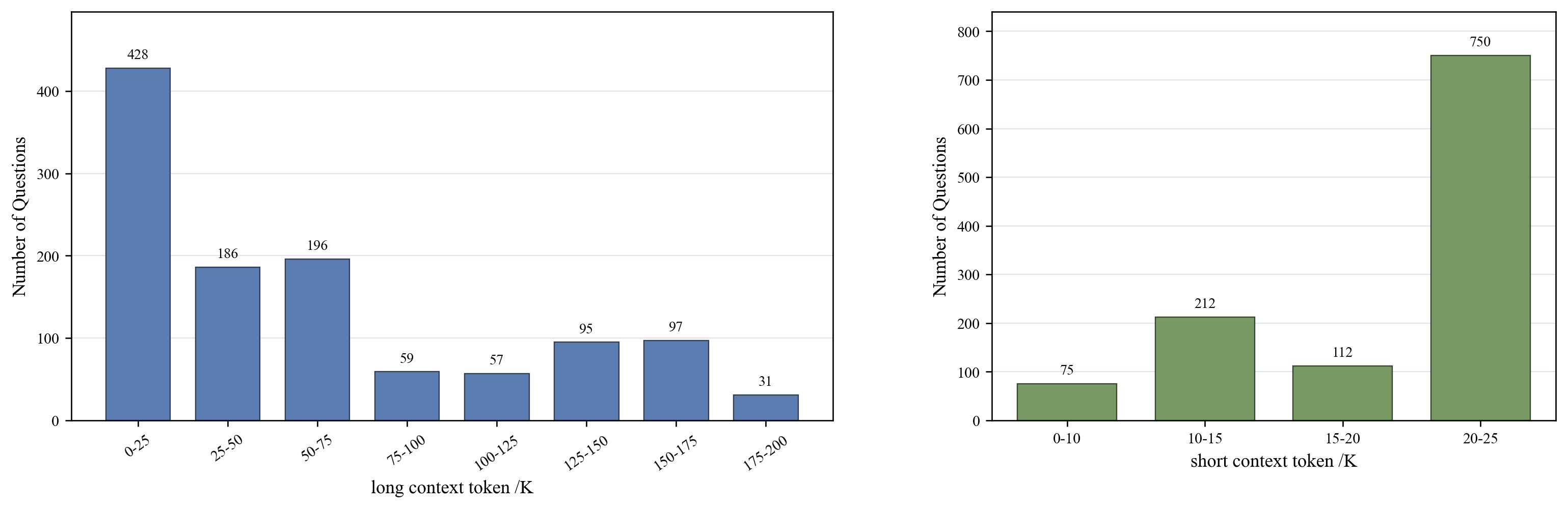}
  \caption{Context-token distribution.}
  \label{fig:context-token-distribution}
\end{figure*}

Figure~\ref{fig:context-token-distribution} shows the context-length distribution of the case-analysis questions. For the long-to-short context conversion, the aggregate token-level compression rate was 54.70\%, while the mean of the per-case compression rates was 25.41\%.

Figure~\ref{fig:system_prompts} reports the system prompts used for these four types of questions mentioned in Section \ref{sec: Experimental Setup}.  Figures~\ref{fig:sample_mcq} -~\ref{fig:sample_rcode_question} present representative examples of the four question types in INS-ActBench.

In quality verification, the process checked: (1) whether each question is fluent and contains sufficient information for deriving the answer; (2) whether the final answer is consistent with the original reference; (3) whether the case background is complete and the simplified version preserves necessary information; (4) whether spreadsheet answer cells are correctly annotated; and (5) whether R-code reference outputs are correct. The average verification time was approximately one minute per INS-Act-Know or INS-Act-Case question and three minutes per INS-Act-Practice question.

\begin{table}[t]
\centering
\small
\begin{tabular}{p{5.5cm}r}
\toprule
\textbf{Selection stage} & \textbf{Percentage} \\
\midrule
Raw questions collected & 100\% \\
Removed because they were more than five years old & 40\% \\
Removed as duplicates or highly similar items & 35\% \\
Removed because images were required & 5\% \\
Removed because answers were incomplete & 5\% \\
Removed because the task format was unsupported & 5\% \\
Final retained questions & 10\% \\
\bottomrule
\end{tabular}
\caption{Approximate question-selection funnel. Percentages are
calculated relative to the initial collection.}
\label{tab:selection_funnel}
\end{table}

\section{Additional Results}
\label{sec:appendixC}

\subsection{Human Expert}
\label{sec:appendixC.1}

Table~\ref{tab:human_expert} presents the individual results of the five human experts, which shows broadly consistent performance on INS-ActBench. Their total scores fall within a narrow range, indicating limited variation across individual experts. Their performance is also relatively stable across the three subsets.

Regarding compensation, the human experts were compensated at a rate equivalent to one day of their regular pay.

\subsection{Different INS-Act-Case Scoring}
\label{sec:appendixC.2}

When we apply a partial-credit rule to the multiple-answer questions in INS-Act-Case, where partially selected correct options receive half credit, all models show a clear score increase compared with the original exact-match scoring scheme. The comparison is reported in Table~\ref{tab:lv2_partial_credit_rescore}. This result indicates that LLMs have already captured part of the relevant information in actuarial cases, but often fail to recover the complete set of correct options. From the perspective of error analysis, the large improvement under partial credit suggests that their case-analysis errors are more often caused by under-selection than by selecting incorrect options. In other words, models tend to miss some necessary case conditions, overlook part of the evidence, or incompletely integrate multiple pieces of case information, rather than systematically misidentifying the meaning of the case materials.

\subsection{More INS-Act-Practice Results}
\label{sec:appendixC.3}

Regarding code efficiency, we have recorded the formula execution time of Python code answers in spreadsheet tasks and the execution time of R code in Docker in R tasks. The differences among models are minimal. Additionally, actuarial science places primary emphasis on numerical correctness. Therefore, efficiency is not used as a core scoring criterion in our benchmark, and is reported only as a reference result, details are shown in Table~\ref{tab:execution_time_statistics}.

\subsection{Question-Weighted Aggregate Score}
\label{app:item_weighted}

The main results use the Subset-weighted Total, calculated as the arithmetic mean of the three subset total scores, so that the three capability dimensions contribute equally despite differences in subset size. For completeness, we additionally report a Question-weighted Total, calculated as the average accuracy over all questions in INS-ActBench. Because INS-Act-Know contains substantially more questions than INS-Act-Case and INS-Act-Practice, the Question-weighted Total is more strongly influenced by knowledge performance and is therefore presented as a supplementary metric. Table~\ref{tab:item_weighted_total} reports both aggregate scores. For the human baseline, the two scores are identical because the human evaluation sampled an equal number of questions from each subset. Under the Question-weighted Total, some LLMs surpass the human experts in overall performance because their generally stronger results on the much larger INS-Act-Know subset contribute disproportionately to the aggregate score.

\section{Data Contamination Diagnostics}
\label{app:contamination}

Because INS-Act-Case and INS-Act-Practice were substantially restructured, we focus the contamination diagnostics on INS-Act-Know. We conduct three complementary analyses using three representative LLMs; the first two analyses are performed on a sampled subset, while the time-split analysis uses the full INS-Act-Know set. The results are reported in Table~\ref{tab:contamination}, suggesting that data contamination has limited impact on INS-Act-Know performance.

\subsection{13-Gram Overlap Analysis}

We randomly sampled approximately 5\% of INS-Act-Know. For each sampled item, we selected three distinctive contiguous 13-word spans and queried them using Google Search. An item was classified as high-exposure when the corresponding official actuarial-association question page or PDF appeared among the top ten results for any of the three queries. All other sampled items were classified as low-exposure. 

Approximately 6.88\% of the sampled items were classified as high-exposure. We separately report model accuracy for the two groups.

\subsection{Model Memorization Test}

For the same sample, each model received only the question and was asked to reconstruct its answer options. For numerical options, we recorded whether the generated numerical value exactly matched the original value. For textual options, we computed an option-level F1 score from 0 to 100.

The models showed limited ability to reconstruct the original option sets. Scores remain in the 20s mainly because models frequently reproduce the correct option while failing to recover the remaining original options.

\subsection{Pre-/Post-Cutoff Time Split}

We consider the actuarial examination questions released in the May 2026 round as post-cutoff items, as they are unlikely to have been included in the training data of the evaluated LLMs. These post-cutoff items account for approximately 4.9\% of INS-Act-Know. We compare their accuracy with the remaining pre-cutoff items to examine whether training-data exposure affects model performance.

\begin{figure*}[t]
\centering
\scriptsize
\renewcommand{\arraystretch}{1.08}
\begin{tabular}{|p{0.47\textwidth}|p{0.47\textwidth}|}
\hline
\textbf{System Prompt: INS-Act-Know} &
\textbf{System Prompt: INS-Act-Case} \\
\hline
\ttfamily
You are an expert actuarial analyst solving the multiple-choice question.\par
Output format rules:\par
- Respond with exactly ONE option letter\par
- Do not output any other text.
&
\ttfamily
You are an expert actuarial analyst solving case-based actuarial multiple-select questions.\par
You will be given case materials and, when relevant, supporting tables that may be useful for solving the question.\par
For each item, use the provided stem as the background and answer the question under that stem.\par
This is a multiple-select question. At least TWO options must be selected.\par
Scoring rule: credit is awarded only if all and only the correct options are selected; any missing option, extra option, incorrect option, or empty response receives zero credit.\par
Output format rules:\par
- Respond with the selected option letters only\par
- Do not output any explanation, reasoning, words, punctuation, or extra text\par
- Write the capital letters together in alphabetical order, for example: AB
\\
\hline
\textbf{System Prompt: Spreadsheet Task of INS-Act-Practice} &
\textbf{System Prompt: R-code Task of INS-Act-Practice} \\
\hline
\ttfamily
You are solving an actuarial spreadsheet task.\par
Write one executable Python code block using openpyxl.\par
Load the input workbook from INPUT\_WORKBOOK, or input.xlsx if the variable is missing.\par
Use the workbook snapshot and task text to understand the sheets, cells, and formulas.\par
Fill the final answer cells in ANSWER\_POSITION; you may use other existing cells for intermediate calculations.\par
Prefer spreadsheet formulas over hard-coded final values when formulas can be built from the workbook data.\par
Do not rename sheets, create extra sheets, or overwrite unrelated content.\par
Save the completed workbook to OUTPUT\_WORKBOOK, or output.xlsx if the variable is missing.\par
Output only the Python code block, with no explanation.
&
\ttfamily
You are solving an actuarial R coding test.\par
Your task is to write accurate R code to compute and solve the given actuarial question.\par
\par
CRITICAL Output Format:\par
- Output ONLY valid, executable R code enclosed in a single R code block.\par
- Do NOT output any standard text explanations, preambles, or concluding remarks.\par
\par
CRITICAL Code Rules:\par
- Assume a fresh base R session. Do not install external packages.\par
- Recreate any inline vectors/matrices exactly.\par
- If datasets like .RData or .csv are provided, assume they are available in the working directory and load them directly.\par
- ALL final numerical answers, statistics, or metrics MUST be explicitly printed to the console so they can be captured by standard output.
\\
\hline
\end{tabular}
\caption{System prompts for the four task types in INS-ActBench.}
\label{fig:system_prompts}
\end{figure*}

\begin{table*}[t]
\centering
\scriptsize
\setlength{\tabcolsep}{4pt}
\renewcommand{\arraystretch}{1.15}
\resizebox{\textwidth}{!}{%
\begin{tabular}{
@{}
l
*{12}{S[table-format=3.2]}
@{}
}
\toprule
\multirow{2}{*}{\textbf{Human}}
& \multicolumn{7}{c}{\textbf{INS-Act-Know}}
& \multicolumn{1}{c}{\textbf{INS-Act-Case}}
& \multicolumn{3}{c}{\textbf{INS-Act-Practice}}
& \multicolumn{1}{c}{\multirow{2}{*}{\textbf{Total}}} \\
\cmidrule(lr){2-8} \cmidrule(lr){9-9} \cmidrule(lr){10-12}
& {\textbf{PS}}
& {\textbf{EF}}
& {\textbf{AMA}}
& {\textbf{LI}}
& {\textbf{NLI}}
& {\textbf{AMO}}
& {\textbf{Total}}
& {\textbf{CA}}
& {\textbf{SS}}
& {\textbf{R}}
& {\textbf{Total}}
& {} \\
\midrule

\textbf{Human Expert 1}
& 94.12 & 88.24 & 100.00 & 88.24 & 94.12 & 86.67 & 92.00
& 82.00
& 85.33 & 94.35 & 89.84
& 87.95 \\

\textbf{Human Expert 2}
& 88.24 & 88.24 & 100.00 & 88.24 & 88.24 & 73.33 & 88.00
& 77.00
& 88.67 & 94.19 & 91.43
& 85.48 \\

\textbf{Human Expert 3}
& 82.35 & 82.35 & 100.00 & 76.47 & 82.35 & 66.67 & 82.00
& 72.00
& 78.45 & 93.91 & 86.18
& 80.06 \\

\textbf{Human Expert 4}
& 70.59 & 70.59 & 100.00 & 64.71 & 70.59 & 40.00 & 70.00
& 81.00
& 70.33 & 82.93 & 76.63
& 75.88 \\

\textbf{Human Expert 5}
& 88.24 & 94.12 & 88.24 & 88.24 & 88.24 & 93.33 & 90.00
& 86.00
& 72.75 & 88.29 & 80.52
& 85.51 \\

\bottomrule
\end{tabular}%
}
\vspace{-1mm}
\caption{Individual human expert performance on INS-ActBench.}
\label{tab:human_expert}
\vspace{-2mm}
\end{table*}

\begin{table*}[t]
\centering
\scriptsize

\begin{minipage}[t]{0.47\textwidth}
\centering
\setlength{\tabcolsep}{4pt}
\renewcommand{\arraystretch}{1.12}
\resizebox{\linewidth}{!}{%
\begin{tabular}{lcc}
\toprule
\textbf{Model}
& \textbf{INS-Act-Case}
& \textbf{\makecell{INS-Act-Case\\(partial credit)}} \\
\midrule

\multicolumn{3}{l}{\textbf{Proprietary LLMs}} \\
\textbf{GPT-5.5}          & \textbf{56.3} & 64.3 \\
\textbf{Claude-Opus-4.7}  & 43.3 & 57.5 \\
\textbf{Gemini-3.1-Pro}   & 55.5 & \textbf{69.5} \\
\textbf{DeepSeek-V4-Pro}  & 43.9 & 58.1 \\
\textbf{Kimi-K2.6}         & 24.8 & 37.7 \\
\textbf{Qwen3.6-Plus}     & 44.6 & 62.1 \\

\midrule
\multicolumn{3}{l}{\textbf{Open-source LLMs}} \\
\textbf{Qwen3.5-35B-A3B}  & 20.2 & 40.0 \\
\textbf{Qwen3-14B}        & 21.0 & 31.2 \\
\textbf{Gemma-3-12B-IT}   & 16.7 & 30.6 \\
\bottomrule
\end{tabular}%
}
\vspace{-1mm}
\caption{Partial-credit re-scoring for INS-Act-Case.}
\label{tab:lv2_partial_credit_rescore}
\end{minipage}
\hfill
\begin{minipage}[t]{0.51\textwidth}
\centering
\setlength{\tabcolsep}{4pt}
\renewcommand{\arraystretch}{1.12}
\resizebox{\linewidth}{!}{%
\begin{tabular}{
@{}
l
*{8}{S[table-format=2.2]}
@{}
}
\toprule
\multirow{2}{*}{\textbf{Model}}
& \multicolumn{4}{c}{\textbf{Spreadsheet}}
& \multicolumn{4}{c}{\textbf{R Code}} \\
\cmidrule(lr){2-5} \cmidrule(lr){6-9}
& {\textbf{Min}}
& {\textbf{Max}}
& {\textbf{Median}}
& {\textbf{Mean}}
& {\textbf{Min}}
& {\textbf{Max}}
& {\textbf{Median}}
& {\textbf{Mean}} \\
\midrule

\multicolumn{9}{l}{\textbf{Proprietary LLMs}} \\
\textbf{GPT-5.5}
& 1.02 & 7.83 & 1.58 & 1.83
& 0.43 & 1.63 & 0.54 & 0.57 \\

\textbf{Claude-Opus-4.7}
& 0.97 & 6.07 & 1.52 & 1.64
& 0.44 & 1.93 & 0.59 & 0.64 \\

\textbf{Gemini-3.1-Pro}
& 1.03 & 9.01 & 1.57 & 1.79
& 0.42 & 3.00 & 0.55 & 0.61 \\

\textbf{DeepSeek-V4-Pro}
& 1.05 & 8.72 & 1.78 & 2.04
& 0.43 & 1.57 & 0.55 & 0.59 \\

\textbf{Kimi-K2.6}
& 1.10 & 3.80 & 1.75 & 1.80
& 0.46 & 2.03 & 0.58 & 0.64 \\

\textbf{Qwen3.6-Plus}
& 0.93 & 10.00 & 1.67 & 1.90
& 0.43 & 3.03 & 0.57 & 0.64 \\

\midrule
\multicolumn{9}{l}{\textbf{Open-source LLMs}} \\
\textbf{Qwen3.5-35B-A3B}
& 0.97 & 7.62 & 1.59 & 1.68
& 0.43 & 1.47 & 0.51 & 0.54 \\

\textbf{Qwen3-14B}
& 1.32 & 4.86 & 2.29 & 2.41
& 0.48 & 2.52 & 0.59 & 0.67 \\

\textbf{Gemma-3-12B-IT}
& 1.25 & 3.56 & 2.01 & 2.06
& 0.45 & 1.40 & 0.55 & 0.59 \\
\bottomrule
\end{tabular}%
}
\vspace{-1mm}
\caption{Execution time statistics for INS-Act-Practice (Unit: second).}
\label{tab:execution_time_statistics}
\end{minipage}

\vspace{-2mm}
\end{table*}

\begin{table*}[t]
\centering
\small
\renewcommand{\arraystretch}{1.1}

\begin{tabular*}{\textwidth}{
@{\extracolsep{\fill}}
l
r
r
r
r
r
@{}
}
\toprule
\textbf{Model}
& \textbf{Know}
& \textbf{Case}
& \textbf{Practice}
& \textbf{Subset-weighted Total}
& \textbf{Question-weighted Total} \\
\midrule

GPT-5.5
& 93.26 & 56.31 & 69.24 & 72.94 & 87.44 \\

Gemini-3.1-Pro
& \textbf{94.53} & 55.53 & 67.91 & 72.66 & \textbf{88.27} \\

DeepSeek-V4-Pro
& 91.25 & 43.86 & 61.67 & 65.59 & 83.63 \\

Claude-Opus-4.7
& 81.10 & 43.34 & 65.38 & 63.27 & 75.99 \\

Qwen3.6-Plus
& 87.83 & 44.56 & 57.43 & 63.27 & 80.27 \\

Kimi-K2.6
& 70.92 & 24.80 & 47.21 & 47.64 & 62.58 \\

Qwen3.5-35B-A3B
& 38.12 & 20.19 & 37.67 & 31.99 & 32.03 \\

Qwen3-14B
& 40.40 & 20.97 & 26.04 & 29.14 & 34.11 \\

Gemma-3-12B-IT
& 34.68 & 16.71 & 20.45 & 23.95 & 29.31 \\

\midrule

Human Baseline
& 84.40
& \textbf{79.60}
& \textbf{84.92}
& \textbf{82.97}
& 82.97 \\

\bottomrule
\end{tabular*}

\caption{Comparison between subset-weighted and question-weighted aggregate scores on INS-ActBench. The Subset-weighted Total assigns equal weight to the Know, Case, and Practice subsets, whereas the Question-weighted Total is calculated over all questions and therefore reflects the subset-size distribution. The best performance in each column is in \textbf{bold}.}
\label{tab:item_weighted_total}
\end{table*}

\begin{table*}[t]
\centering
\small
\setlength{\tabcolsep}{5pt}
\renewcommand{\arraystretch}{1.15}
\resizebox{\textwidth}{!}{%
\begin{tabular}{
@{}
l
*{10}{c}
@{}
}
\toprule
& & \multicolumn{3}{c}{\textbf{13-gram overlap}}
& \multicolumn{3}{c}{\textbf{Memorization}}
& \multicolumn{3}{c}{\textbf{Time split}} \\
\cmidrule(lr){3-5}
\cmidrule(lr){6-8}
\cmidrule(lr){9-11}
\textbf{Model}
& \textbf{Full}
& \textbf{High}
& \textbf{Low}
& \textbf{Gap}
& \textbf{High}
& \textbf{Low}
& \textbf{Gap}
& \textbf{Post}
& \textbf{Pre}
& \textbf{Gap} \\
\midrule

GPT-5.5
& 93.26
& 94.12
& 91.30
& \textcolor{green!50!black}{+2.81}
& 28.98
& 26.87
& \textcolor{green!50!black}{+2.11}
& 94.46
& 93.20
& \textcolor{green!50!black}{+1.26} \\

Claude-Opus-4.7
& 81.10
& 82.35
& 84.35
& \textcolor{red}{-1.99}
& 27.21
& 29.51
& \textcolor{red}{-2.29}
& 79.88
& 81.16
& \textcolor{red}{-1.28} \\

Gemini-3.1-Pro
& 94.53
& 94.12
& 95.22
& \textcolor{red}{-1.10}
& 37.95
& 32.47
& \textcolor{green!50!black}{+5.48}
& 95.69
& 94.47
& \textcolor{green!50!black}{+1.22} \\

\bottomrule
\end{tabular}%
}
\vspace{-1mm}
\caption{Data-contamination diagnostics on INS-Act-Know. Full denotes each model's original accuracy on the complete INS-Act-Know subset. In the 13-gram overlap and memorization analyses, High and Low denote results on the high-exposure and low-exposure groups. In the time-split analysis, Post denotes accuracy on questions released in the May 2026 examination round, while Pre denotes accuracy on the remaining earlier questions. Gap is calculated as High minus Low for the first two analyses and Post minus Pre for the time-split analysis. Gaps are computed from unrounded scores and may therefore differ slightly from subtraction using the displayed values.}
\label{tab:contamination}
\vspace{-2mm}
\end{table*}

\begin{figure*}[t]
    \centering
    \begin{tcolorbox}[colback=white, colframe=black, boxrule=1pt, arc=2pt, left=6pt, right=6pt, top=6pt, bottom=6pt]
        
        \centering \textbf{Sample Data Instance: Annotated Multiple-Choice Question}
        \vspace{0.5em}
        \flushleft
        
        \textbf{Annotation:} \\
        \vspace{0.3em}
        \small
        \begin{tabular}{|l|l|l|l|l|}
            \hline
            \textbf{Dataset:} INS-Act-Know & \textbf{Datasubset:} Life Insurance (LI) & \textbf{Source:} SOA & \textbf{Numerical:} yes & \textbf{id:} 58 \\
            \hline
        \end{tabular}
        \vspace{0.8em}
        \normalsize
        
        \textbf{Question:} \\
        For an annuity-due that pays 100 at the beginning of each year that (45) is alive, you are given:
        \begin{enumerate}[label=(\roman*), noitemsep, topsep=2pt]
            \item Mortality for standard lives follows the Standard Ultimate Life Table.
            \item The force of mortality for standard lives age $45 + t$ is represented as $\mu_{45+t}^{SULT}$.
            \item The force of mortality for substandard lives age $45 + t$, $\mu_{45+t}^{S}$, is defined as:
            \[
            \mu_{45+t}^{S} = 
            \begin{cases}
            \mu_{45+t}^{SULT} + 0.05, & \text{for } 0 \leq t < 1 \\
            \mu_{45+t}^{SULT}, & \text{for } t \geq 1
            \end{cases}
            \]
            \item $i = 0.05$
        \end{enumerate}
        Calculate the actuarial present value of this annuity for a substandard life age 45.
        
        \vspace{0.8em}
        
        \textbf{Options:} (A) 1700 \quad (B) 1710 \quad (C) 1720 \quad (D) 1730 \quad (E) 1740
        
        \vspace{0.5em}
        \hrule
        \vspace{0.5em}
        
        \textbf{Answer:} \texttt{A}
        
    \end{tcolorbox}
    \caption{A representative example of a Life Insurance question from the INS-Act-Know subset.}
    \label{fig:sample_mcq}
\end{figure*}

\begin{figure*}[t]
    \centering
    \begin{tcolorbox}[colback=white, colframe=black, boxrule=1pt, arc=2pt, left=6pt, right=6pt, top=6pt, bottom=6pt]
        
        \centering \textbf{Sample Data Instance: Case-based Multiple-select Question}
        \vspace{0.5em}
        \flushleft
        
        \textbf{Metadata Annotation:} \\
        \vspace{0.3em}
        \small
        \begin{tabular}{|l|l|l|}
            \hline
            \textbf{Dataset:} INS-Act-Case & \textbf{Source:} SOA  & \textbf{id:} 113 \\
            \hline
        \end{tabular}
        \vspace{0.8em}
        \normalsize
        
        \textbf{Case-long:} \\
        1.1    RPPC Corporation History
        
        RPPC was established in 2005 with head offices in Luxembourg by four founding partners...(132K)
        \vspace{0.8em}
        
        \textbf{Case-short:} \\
        1.1 RPPC Corporation History
        
        The business roots began with the coffee shop, owned by the Ruiz family since 1995...(24K)
        \vspace{0.8em}
        
        \textbf{Stem:} \\
        You are engaged by Blue Jay Air (BJA) to provide advice on its international expansion strategy. Since being separated from RPPC, BJA has continued pursuing ambitious growth opportunities. (Case Study Section 2.7) Research by your team shows that the median debt-to-equity ratio in the airline industry is 5.5x.
        
        The BJA Finance Team updated the analysis of two alternatives for the international expansion project: one is to purchase the international plane fleet, and the other is to upgrade the existing plane fleet for international travel. This capital budgeting model is in tab Q2\_b and uses the companywide cost of capital and expected free cash flows to calculate net present value. BJA needs an independent review of this model in order to obtain a fixed debt financing agreement.
        
        \vspace{0.8em}
        
        \textbf{Question:} \\
        Based on the Excel file and the financial statements, which of the following statements about BJA's net debt-to-value ratio and capital structure evolution over the past three years are correct?
        
        \vspace{0.8em}
        
        \textbf{Options:}
        \begin{itemize}[noitemsep, topsep=0pt, leftmargin=2em]
            \item[(A)] the 2020 net debt-to-value ratio was approximately 96\%
            \item[(B)] the 2021 net debt-to-value ratio was approximately 86\%
            \item[(C)] the 2022 net debt-to-value ratio was approximately 64\%
            \item[(D)] BJA's leverage decreased over the past three years
            \item[(E)] BJA's debt-to-equity ratio has fallen below the industry median of 5.5x
            \item[(F)] retained earnings over the past three years were all converted into cash dividends, thereby depressing equity
            \item[(G)] net debt in the past three years was 349, 469, and 538 respectively
            \item[(H)] shareholder equity in the past three years was 199, 89, and 22 respectively
        \end{itemize}
        
        \vspace{0.5em}
        \hrule
        \vspace{0.5em}
        
        \textbf{Answer:} \texttt{ABCDEG}
        
    \end{tcolorbox}
    \caption{A representative example of a case-based multiple-select question from the INS-Act-Case subset.}
    \label{fig:sample_case}
\end{figure*}

\begin{figure*}[t]
    \centering
    \begin{tcolorbox}[colback=white, colframe=black, boxrule=1pt, arc=2pt, left=6pt, right=6pt, top=6pt, bottom=6pt]
        
        \centering \textbf{Sample Data Instance: Spreadsheet Question}
        \vspace{0.5em}
        \flushleft
        
        \textbf{Metadata Annotation:} \\
        \vspace{0.3em}
        \small
        \begin{tabular}{|l|l|l|l|}
            \hline
            \textbf{Dataset:} INS-Act-Practice & \textbf{Datasubset:} Spreadsheet & \textbf{Source:} IFoA & \textbf{id:} 2 \\
            \hline
        \end{tabular}
        \vspace{0.8em}
        \normalsize
        
        \textbf{Question:} \\
        Calculate the annual effective inflation rate over the previous 12 months for each month from January 2004 to December 2019 using the index values provided.
        
        \vspace{0.8em}
        
        \textbf{Template:} See Figure~\ref{fig:spreadsheet_template}.           \textbf{Answer Position:} \texttt{'G2 (i)'!D17:D208}
        
        \vspace{0.8em}
        
        
        \vspace{0.5em}
        \hrule
        \vspace{0.5em}
        
        \textbf{Answer:} See Figure~\ref{fig:spreadsheet_answer}.
        
    \end{tcolorbox}
    \caption{A representative example of a spreadsheet question from the INS-Act-Practice subset.}
    \label{fig:sample_spreadsheet_question}
\end{figure*}

\begin{figure*}[t]
    \centering
    \begin{tcolorbox}[colback=white, colframe=black, boxrule=1pt, arc=2pt, left=6pt, right=6pt, top=6pt, bottom=6pt]
        
        \centering \textbf{Sample Data Instance: Annotated R-Code Question}
        \vspace{0.5em}
        \flushleft
        
        \textbf{Metadata Annotation:} \\
        \vspace{0.3em}
        \small
        \begin{tabular}{|l|l|l|l|}
            \hline
            \textbf{Dataset:} INS-Act-Practice & \textbf{Datasubset:} R-Code & \textbf{Source:} IFoA & \textbf{id:} 39 \\
            \hline
        \end{tabular}
        \vspace{0.8em}
        \normalsize
        
        \textbf{Question:} \\
        An analysis was carried out to investigate the fairness of two exam markers. They both marked the same 150 exam papers, with 10 questions and total possible marks of 100 for each exam paper. The data were collected and arranged into 10 equally spaced groups, with marks rounded to the nearest whole number.
        
        \vspace{0.5em}
        Below are the frequencies of the marks given by each of the exam markers:
        
        \vspace{0.4em}
        \centering
        \small
        \setlength{\tabcolsep}{5pt}
        \begin{tabular}{|l|c|c|c|c|c|c|c|c|c|c|}
            \hline
            \textbf{Exam marks} & \textbf{0--10} & \textbf{11--20} & \textbf{21--30} & \textbf{31--40} & \textbf{41--50} & \textbf{51--60} & \textbf{61--70} & \textbf{71--80} & \textbf{81--90} & \textbf{91--100} \\
            \hline
            \textbf{Marker 1} & 1 & 8 & 14 & 22 & 33 & 34 & 21 & 9 & 6 & 2 \\
            \hline
            \textbf{Marker 2} & 0 & 4 & 16 & 25 & 27 & 42 & 23 & 4 & 9 & 0 \\
            \hline
        \end{tabular}
        
        \vspace{0.7em}
        \flushleft
        One of the marked exam papers is selected at random and the scores given by each of the markers are analysed further by question. Below are the scores given by each of the exam markers for this selected exam paper for each of the 10 questions:
        
        \vspace{0.4em}
        \centering
        \small
        \setlength{\tabcolsep}{8pt}
        \begin{tabular}{|l|c|c|c|c|c|c|c|c|c|c|}
            \hline
            \textbf{Question} & \textbf{1} & \textbf{2} & \textbf{3} & \textbf{4} & \textbf{5} & \textbf{6} & \textbf{7} & \textbf{8} & \textbf{9} & \textbf{10} \\
            \hline
            \textbf{Marker 1} & 4 & 1 & 5 & 1 & 4 & 6 & 4 & 5 & 3 & 6 \\
            \hline
            \textbf{Marker 2} & 3 & 2 & 4 & 0 & 3 & 4 & 2 & 3 & 3 & 6 \\
            \hline
        \end{tabular}
        
        \vspace{0.7em}
        \flushleft
        Perform a suitable test to determine whether the difference in the mean scores of the two markers is zero or not, at the 5\% confidence level, taking into account that the two markers have marked the same exam paper.
        
        \vspace{0.5em}
        \hrule
        \vspace{0.5em}
        
        \textbf{Ground Truth (Reference Output):} \texttt{statistic = 2.86199428; p\_value = 0.01871942}

        \vspace{0.3em}

        \textbf{Scoring:} \texttt{[2.861994, 0.018719]}
        
    \end{tcolorbox}
    \caption{A representative example of an R-code question from the INS-Act-Practice subset.}
    \label{fig:sample_rcode_question}
\end{figure*}

\begin{figure*}[t]
    \centering

    \begin{minipage}[t]{0.48\textwidth}
        \centering
        \includegraphics[height=0.6\textheight,keepaspectratio]{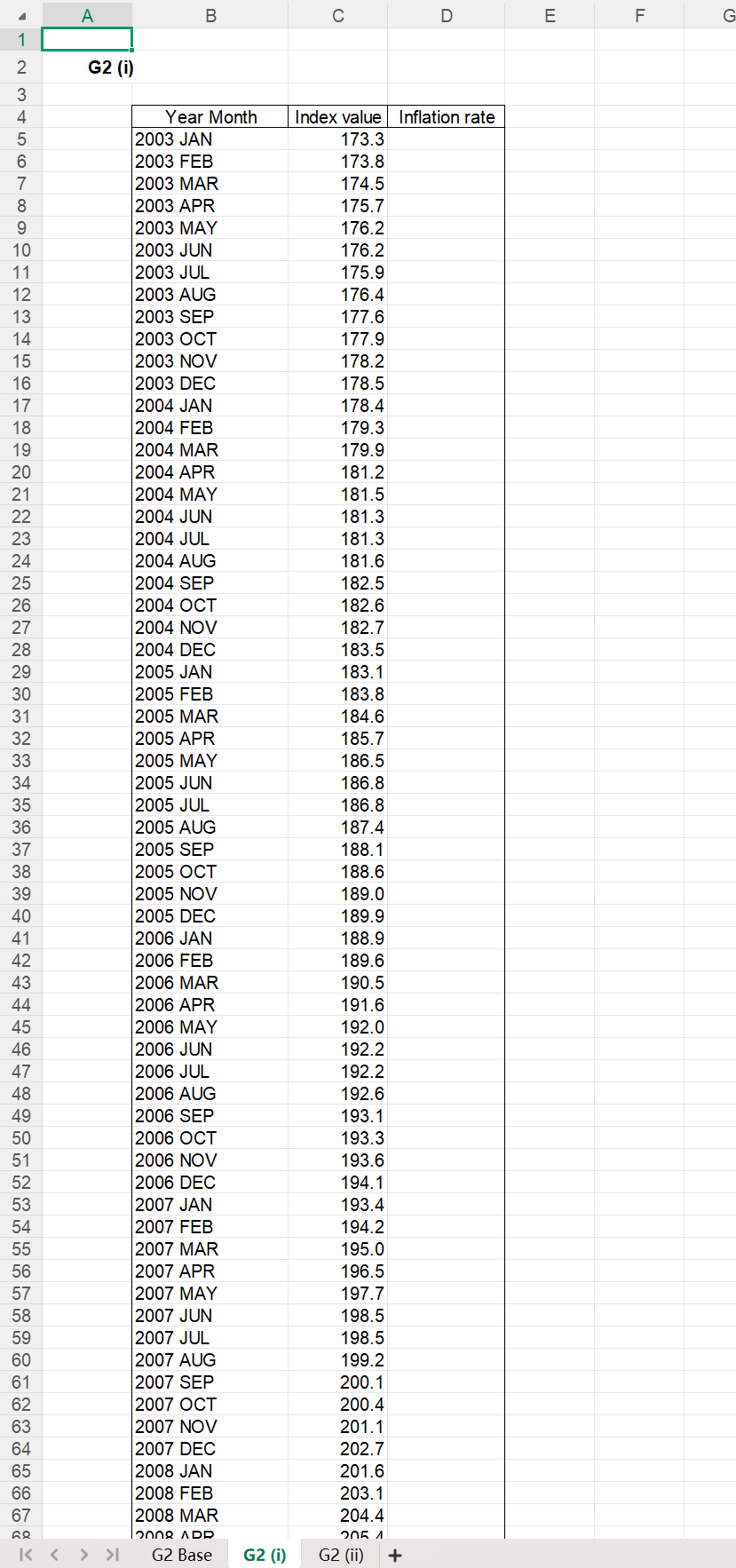}
        \caption{Template workbook example.}
        \label{fig:spreadsheet_template}
    \end{minipage}
    \hfill
    \begin{minipage}[t]{0.48\textwidth}
        \centering
        \includegraphics[height=0.6\textheight,keepaspectratio]{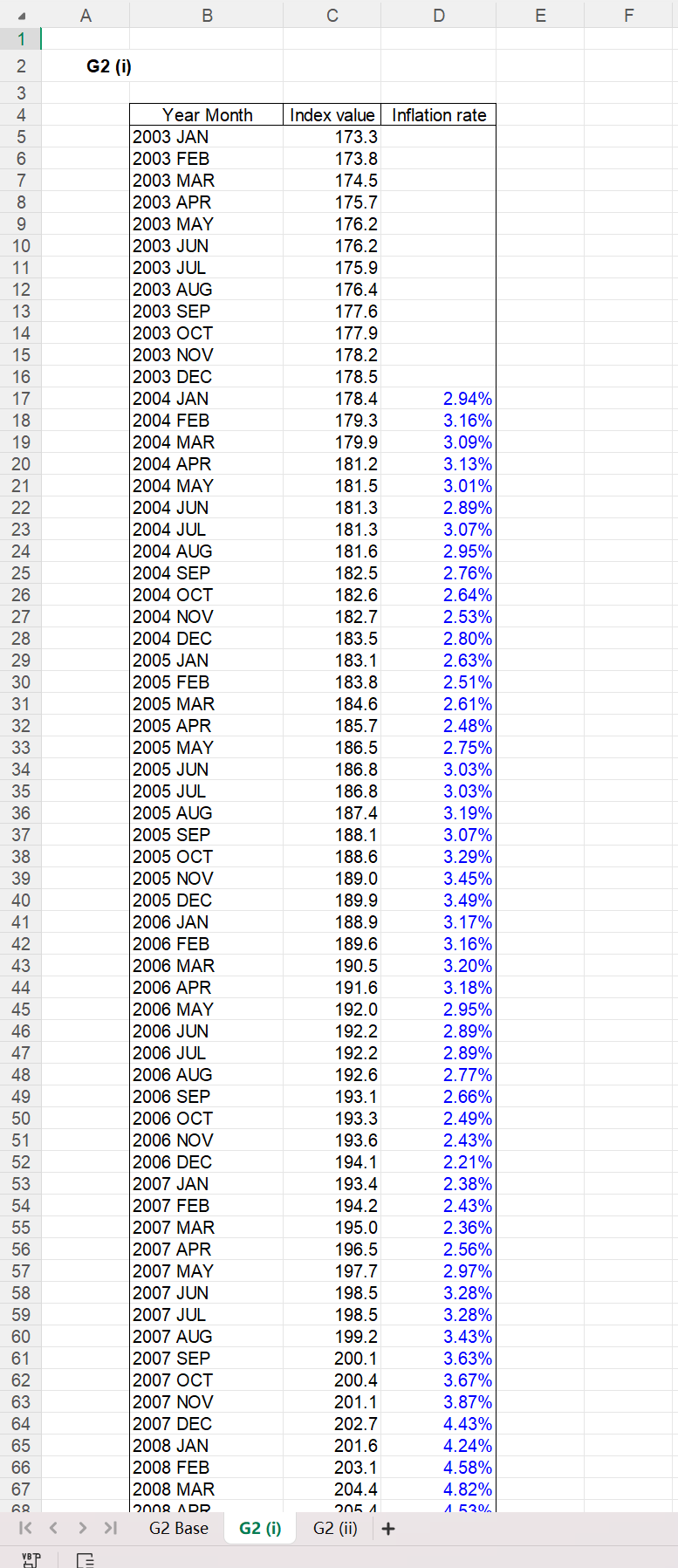}
        \caption{Ground-truth answer example.}
        \label{fig:spreadsheet_answer}
    \end{minipage}

\end{figure*}

\end{document}